\documentclass[conference]{IEEEtran}
\IEEEoverridecommandlockouts
\usepackage{cite}
\usepackage{amsmath,amssymb,amsfonts}
\usepackage{algorithmic}
\usepackage{graphicx}
\usepackage{textcomp}
\usepackage{subcaption}
\usepackage{multirow}
\usepackage{bbm}
\usepackage{xcolor}
\usepackage{url}

\def\BibTeX{{\rm B\kern-.05em{\sc i\kern-.025em b}\kern-.08em
    T\kern-.1667em\lower.7ex\hbox{E}\kern-.125emX}}
\begin{document}

\title{Contrastive Learning of\\ Generalized Game Representations \\
\thanks{This project has received funding from the EU’s Horizon 2020 programme under grant agreement No 951911.}
}

\author{\IEEEauthorblockN{Chintan Trivedi}
\IEEEauthorblockA{\textit{Institute of Digital Games} \\
\textit{University of Malta}\\
Msida, Malta \\
ctriv01@um.edu.mt}
\and
\IEEEauthorblockN{Antonios Liapis}
\IEEEauthorblockA{\textit{Institute of Digital Games} \\
\textit{University of Malta}\\
Msida, Malta \\
antonios.liapis@um.edu.mt}
\and
\IEEEauthorblockN{Georgios N. Yannakakis}
\IEEEauthorblockA{\textit{Institute of Digital Games} \\
\textit{University of Malta}\\
Msida, Malta \\
georgios.yannakakis@um.edu.mt}
}
\maketitle

\begin{abstract}
Representing games through their pixels offers a promising approach for building general-purpose and versatile game models. While games are not merely images, neural network models trained on game pixels often capture differences of the visual style of the image rather than the content of the game. As a result, such models cannot generalize well even within similar games of the same genre. In this paper we build on recent advances in contrastive learning and showcase its benefits for representation learning in games. Learning to contrast images of games not only classifies games in a more efficient manner; it also yields models that separate games in a more meaningful fashion by ignoring the visual style and focusing, instead, on their content. Our results in a large dataset of sports video games containing 100k images across 175 games and 10 game genres suggest that contrastive learning is better suited for learning generalized game representations compared to conventional supervised learning. The findings of this study bring us closer to universal visual encoders for games that can be reused across previously unseen games without requiring retraining or fine-tuning.
\end{abstract}


\section{Introduction}\label{sec:intro}

The use of pixels to represent games is gradually dominating the field of artificial intelligence (AI) in games \cite{yannakakis2018artificial} with applications that vary from gameplaying agents \cite{mnih2013playing,ha2018world,badia2020agent57}, and game content generation \cite{kim2020learning,liu2020deep} all the way to player affect modeling \cite{makantasis2019pixels,makantasis2021pixels}. Deep learning methods---predominately variants of convolutional neural networks (ConvNets)---process the RGB pixels of the game and convert them into a compressed representation that approximates the internal state of the game world. While computer vision methods appear to offer certain capacities when it comes to their general use across games, they come with certain limitations including the computational cost of training and their poor reusability across games.

\begin{figure}[t]
\centering
\includegraphics[width=\columnwidth]{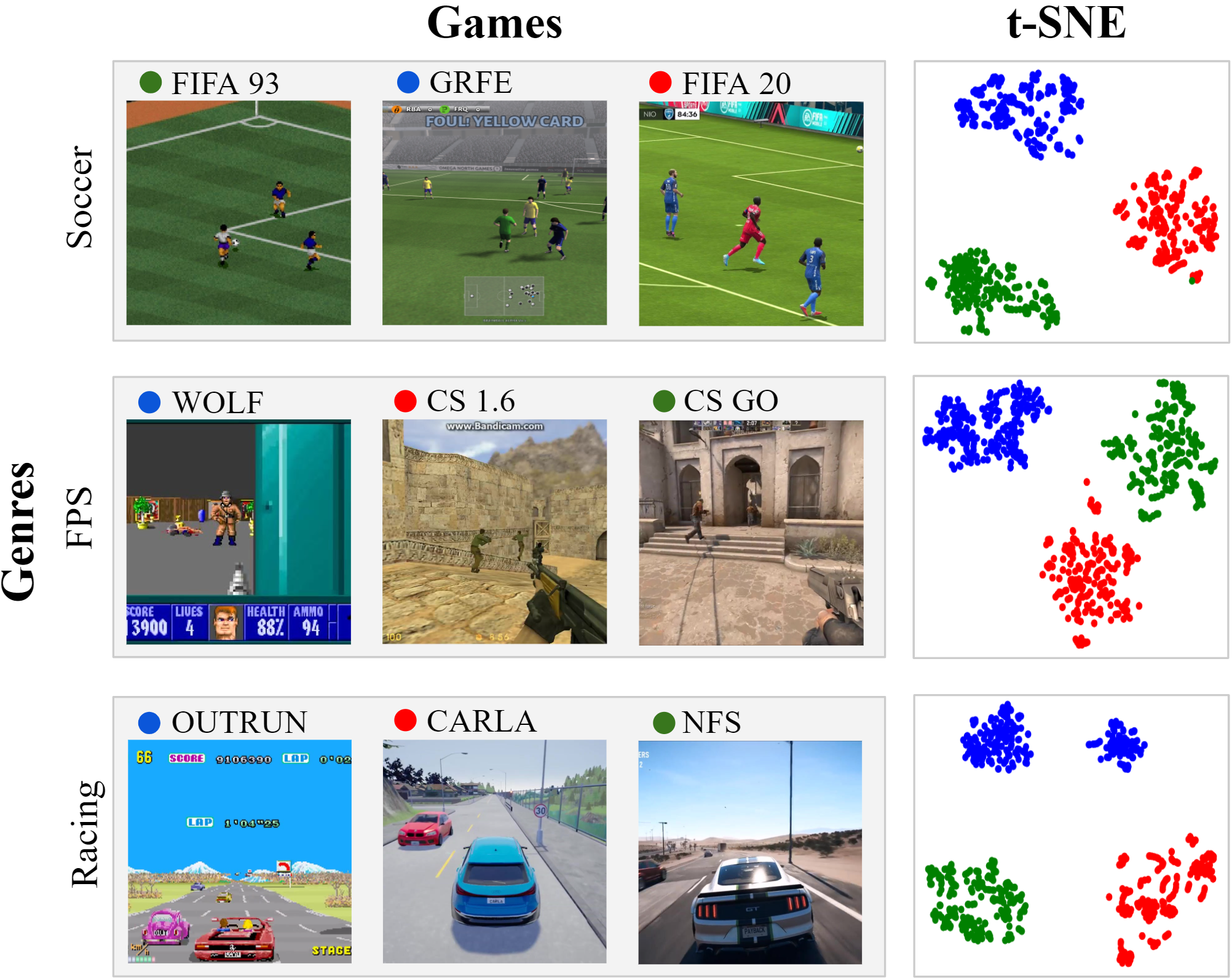}
\caption{Examples of the \emph{domain gap} problem observed when comparing different games of the same genre. The scatter-plots (right) highlight the intra-genre domain discrepancies with the help of t-SNE visualization of Imagenet-ResNet50 feature vectors on screenshots of various games (left).}
\label{fig:domaingapexample}
\end{figure}

A common approach for reducing computation effort and improving on generality is to use and fine-tune pretrained models such as ResNet \cite{he2016deep} that are trained on large datasets such as ImageNet \cite{russakovsky2015imagenet}. While such models can detect many common everyday real-world objects, they still require fine-tuning when applied to games as they are far from ideal replicas of the real-world. Importantly, games are not merely image representations; gameplay images contain both functional properties associated with the game genre (e.g. corridors, tracks and platforms that define movement constraints) and aesthetic elements unique to each game (e.g. the various art styles available in match-3 tile games). As a visual example of this issue, \figurename{} \ref{fig:domaingapexample} displays games with similar \emph{content} (i.e. same game genre) but with different visual \emph{style}. The representations obtained from these game images with a pre-trained ResNet-50 model can be visualized on a 2D plane using $t$-distributed stochastic neighbor embedding (t-SNE) \cite{maaten2008visualizing}. We observe that the 2D embeddings of representations of 3 different Soccer, FPS and Racing games form their own separate ``clusters'' in the Euclidean space. This phenomenon, named \emph{domain gap} \cite{wang2018deep}, occurs due to the visual styling differences in each game, leading to shifts in the distributions. It is therefore expected that any AI algorithm that builds on representations obtained from a pre-trained model will only operate well on the particular game it is trained on, and will require substantial fine-tuning (retraining or transfer learning) in order to be of use on other games. This lack of generalizability and reusability makes pixel-based deep learning impractical across different games even if they belong to the same genre. 

To tackle the above-mentioned challenges in this paper we introduce \emph{contrastive learning} \cite{chen2020simple} as a novel way to approach the domain gap challenge in games. Our hypothesis is that by contrasting pixel-based representations---instead of merely classifying them---we can fine-tune pre-trained ConvNet models that better capture the underlying content of the game rather than its style. 
We test our hypothesis on a new dataset, namely \emph{Sports10}, featuring 100k images of 175 different games across 10 sports game genres. By training ConvNets on this dataset via fully supervised and contrastive learning techniques, we show that the latter is better suited for not only achieving higher genre-classification accuracy, but more importantly, for attaining better generalization capacity. Findings suggest that contrastive learning yields more general pixel-based representations of games by focusing more on the content of the game while demonstrating better invariance to the visual styling differences in the provided images.

\section{Representation Learning in Games}\label{sec:bakcground}

Recent research in machine learning has often developed intelligent systems that use pixel information as the (only) input. In games, pixel representations have been used for content-based retrieval for moments \cite{zhang2018crawling} and to predict players' affective states \cite{makantasis2019pixels}; however, the most prominent application is Deep Reinforcement Learning. Mnih \emph{et al.} \cite{mnih2013playing} introduced one of the first game-playing agents for Atari games that learns control-policy directly from pixels using ConvNets. This work was extended by Kempka \emph{et al.} \cite{kempka2016vizdoom} to play \emph{Doom} (id Software, 1993) using screen buffer and depth information processed by ConvNets in the ViZDoom platform. Ha and Schimdhuber \cite{ha2018world} presented a recurrent model that uses convolutional auto-encoders with temporal memory to create a \emph{world model} of the game. Since auto-encoders are designed to reconstruct the input from its encoding, this world model can be seen as a compressed representation of the game that encodes both \emph{content} and \emph{style} information of the environment. 

A number of approaches attempt to encode only the \emph{content} information so that it makes the subsequent policy-learning task easier. Srinivas \emph{et al.} \cite{srinivas2020curl}, for instance, combined policy learning along with representation learning in a unified framework which yields representations that contain only the \emph{content} information of the game, which is a better resemblance of a game's internal state. Another direction focusing on the separation of \emph{content} from \emph{style} is to derive style-invariant representations of the game environment using data augmentation techniques---such as color shift, gray-scale conversion, etc.---that produce different styles of the same image \cite{laskin2020reinforcement}. Such learning frameworks encourage the convolutional encoder to ignore style-related information of the game that is present in the screen pixels and focus more on the content. The scope of generalization in such approaches, however, still remains limited to the game environment that the visual encoder is being trained on. As a result, these methods are still susceptible to the domain gap problem described earlier.

When it comes to video games, Luo \emph{et al.} \cite{luo2018player} show how to use transfer learning to train ConvNets for extracting game events, but they do not focus on generalization. Khameneh and Guzdial \cite{khameneh2020entity} try to tackle generalization, but their method operates on game events extracted from internal state rather than pixels. To the best of our knowledge there have been no attempts to specifically tackle the \emph{domain gap} challenge in computer vision for video games. Motivated by this knowledge gap and inspired by recent trends in computer vision \cite{mitrovic2020representation} we test the capacity of contrastive learning \cite{chen2020simple, khosla2020supervised} to train visual encoders for video games and evaluate the extent to which it can mitigate domain gap problems by learning representations that can generalize over different games.

\section{Generalization}
\label{sec:generalization} 

While the broader definition of generalization in AI may be rather subjective and open-ended \cite{kawaguchi2017generalization}, in this section we provide its formal definition for our work and restrict its scope in terms of representation learning in video games. We then discuss how to quantitatively measure it in order to evaluate the performance of our generalization models.

\subsection{Definition}

\begin{figure}[!tb]
    \centering
    \includegraphics[width=\linewidth]{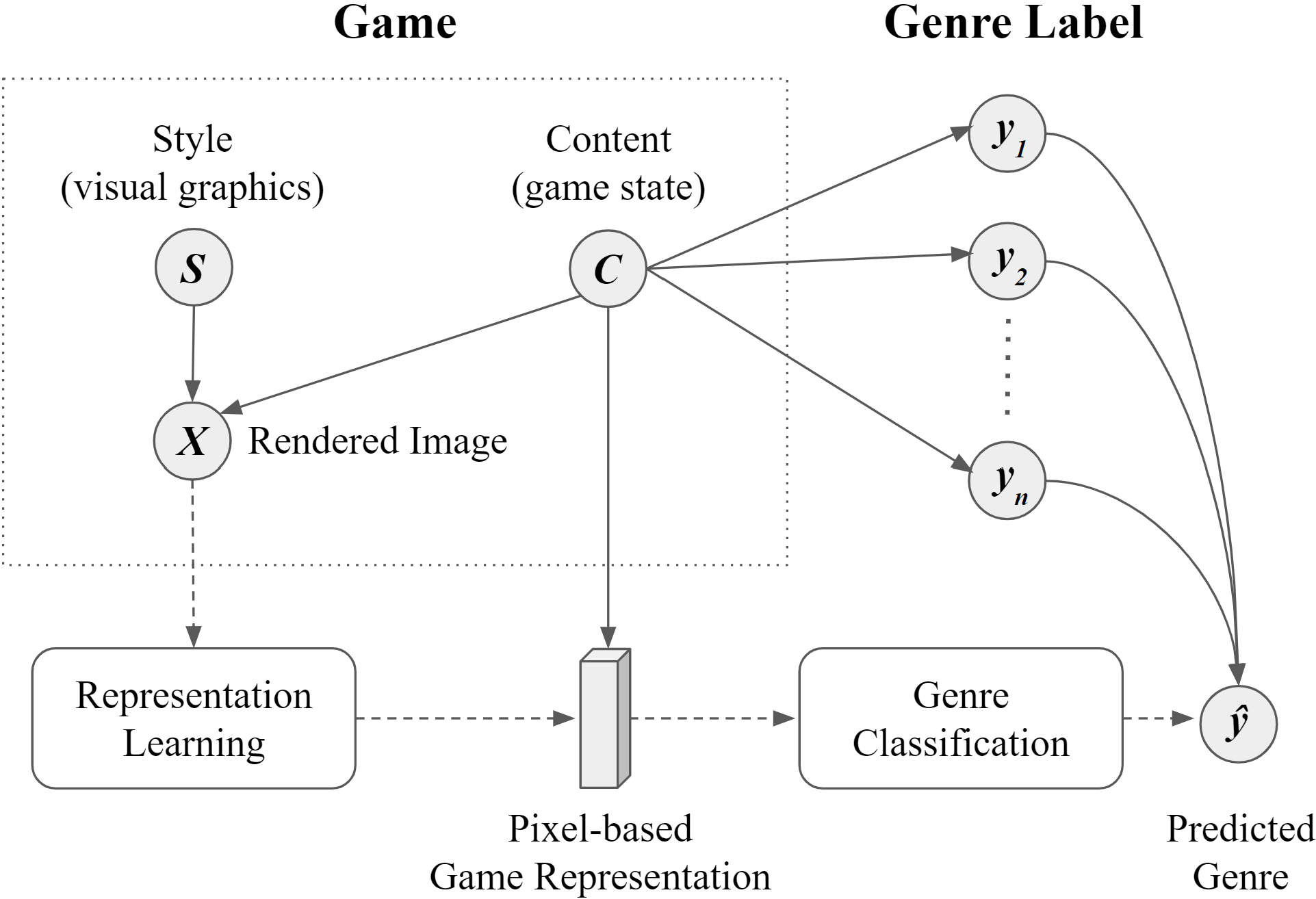}
    \caption{The causality framework for generalization in games. Solid lines represent a causal dependency and dashed lines represent flow of data. For a given game belonging to one of the genres ($y_{1..n}$), the graph showcases the relationship of style ($S$) and content ($C$) of that game's image ($X$) with its learned representation and the predicted genre category ($\hat{y}$). }
    \label{fig:causalityframework}
\end{figure} 

\begin{figure*}[!tb]
\centerline{\includegraphics[width=\textwidth]{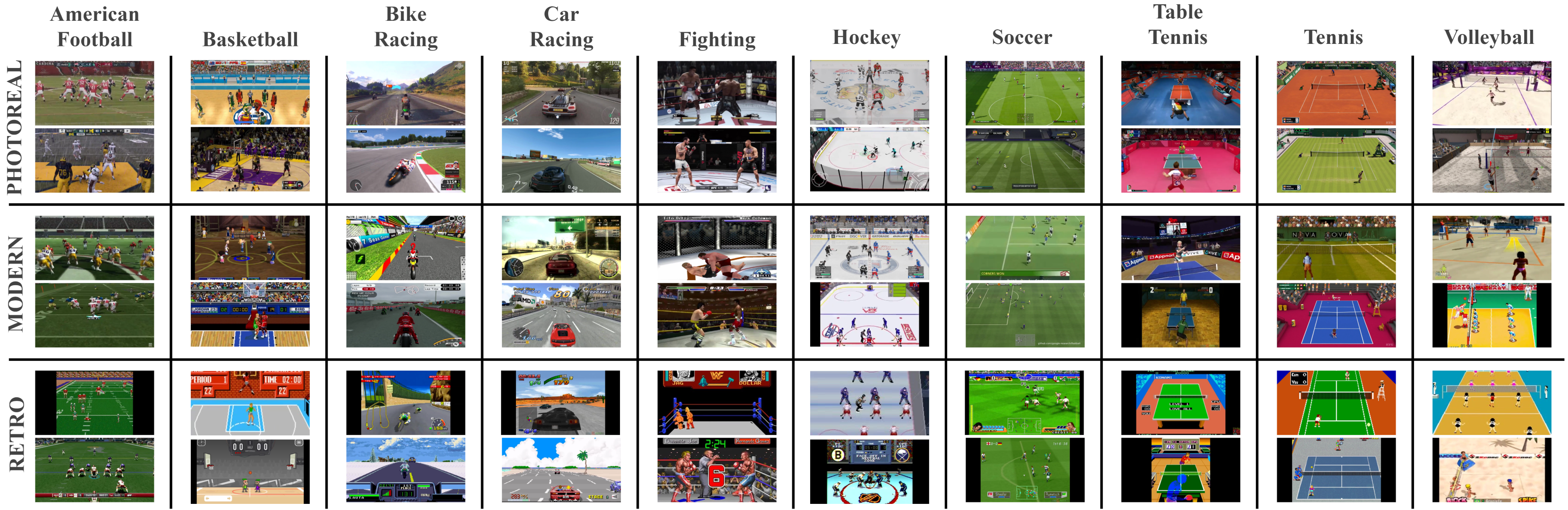}}
\caption{A subset of the \emph{Sports10} Dataset showcasing the variation in graphic styling of different games across each of the 10 sports genres selected.}
\label{dataset}
\end{figure*}

In this paper we define \emph{generalization} as the ability of a trained ConvNet model to process pixels of a game screen and extract a meaningful representation of the game's content without being affected by the graphic styling of the game. Figure \ref{fig:causalityframework} visualizes generalisation in the form of a causality graph \cite{mitrovic2020representation} showing the style-invariance requirement for a generalized game representation. Under this causality framework, only the content of the game defines the game genre while the style only affects the rendered image of the game. Different video games belonging to the same genre can be thought of as having the same \emph{content} but varying \emph{style}. Thus, a model that generalizes well should be able to extract representations from any game within the same genre without exhibiting a domain gap.

\subsection{Measuring Generalization}

We propose to measure generalization in terms of the mitigation of domain gap across different styles of games of the same genre. This can be measured in the representation space, which is a $d$-dimensional Euclidean Space where $d$ is the size of the latent representation. In this formulation, better generalization means the representations of different games of the same genre form a compact cluster and are well-separated from representation clusters of other genres. Hence, we choose the \emph{Silhouette Score} \cite{rousseeuw1987silhouette} as a metric for evaluating the quality of representation clusters.

Assume that a given dataset $\mathcal D$ contains images $\mathcal X$ belonging to a set of game genres $\mathcal Y = \{y_1, y_2, ..., y_n\}$. Let's denote the $i^{th}$ image of this dataset as $X_i \in \mathcal X$ having genre label $y_i \in \mathcal Y$. The pixel-representation extracted from $X_i$ is denoted by $x_i$ where $|x_i|=d$. Let's denote the average intra-cluster Euclidean distance (cluster compactness) of this image representation within its own genre as $a(x_i)$. Likewise, let's denote the average inter-cluster distance (separation from other clusters) to the nearest cluster as $b(x_i)$. Then, the Silhouette Coefficient $s(X_i)$ of this image is defined as:

\begin{equation}
s(X_i) = \frac{b(x_i) - a(x_i)}{max\{a(x_i),b(x_i)\}}
\end{equation}

\noindent with the assumption that the set of images belong to more than one game genre ($|\mathcal Y|>1$). The combined Silhouette Score $S(\mathcal D)$ for the entire set of images can be defined as the average silhouette coefficient over all points $i$ in $\mathcal D$. This $S(\mathcal D)$ lies within $[-1,1]$ with higher values indicating more compact and better separated clusters. If we group visually different games of the same genre and assign them the same cluster label, we expect the silhouette score of this group of images would quantify domain gap issues present in the dataset. Based on the properties of $S(\mathcal D)$, we argue that it is a good measurement of generalization capability across different games.

In addition to the silhouette score, we use \emph{t-SNE} \cite{maaten2008visualizing} for qualitative analysis by visualizing the representation space, which offers an approximate 2-dimensional projection of the $d$-dimensional representations learned by the visual encoder. Note that this technique is merely used in our work for its intuitive visualization of what the ConvNet models have learned, rather than evaluation or comparison of the results.

\section{Sports10: A Dataset of Sports Game Genres}
\label{sec:dataset}

In order to learn style-invariant features of games from pixels, we created a new game genres dataset\footnote{Available at github.com/ChintanTrivedi/contrastive-game-representations} for our experiments. It contains $100,000$ gameplay images of 175 different games across 10 game genres that are extracted from publicly available sports video game titles. Each of the 10 genres---\textit{American Football, Basketball, Bike Racing, Car Racing, Fighting, Hockey, Soccer, Table Tennis, Tennis} and \textit{Volleyball}---contain exactly $10,000$ hand-curated images of the game-play sequence, ensuring the removal of all menu, transitions or cutscenes of the game. \figurename{} \ref{dataset} shows a glimpse of the variety of games that are a part of this dataset, divided into our interpretation of three visual styling categories: \emph{retro} (arcade-style, 1990s and earlier), \emph{modern} (roughly 2000s) and \emph{photoreal} (roughly late 2010s). The genre and styling-wise breakdown of total games in our dataset is given in Table \ref{datasetinfo}.

\begin{table}[!tb]
\begin{center}
\caption{Summary of total games per genre in the dataset and their distribution across the different visual styling categories.}
\label{datasetinfo}
\begin{tabular}{|l|c|c|c|c|}
\hline
\multicolumn{1}{|c|}{\textbf{Game Genre}} & \textbf{Retro} & \textbf{Modern} & \textbf{Photoreal} & \textbf{Total} \\ \hline
American Football & 2 & 11 & 6 & \textbf{19} \\ 
Basketball & 3 & 12 & 3 & \textbf{18} \\ 
Bike Racing & 8 & 7 & 4 & \textbf{19} \\ 
Car Racing & 5 & 5 & 5 & \textbf{15} \\ 
Fighting & 3 & 11 & 9 & \textbf{23} \\ 
Hockey & 9 & 7 & 1 & \textbf{17} \\ 
Soccer & 7 & 8 & 2 & \textbf{17} \\ 
Table Tennis & 3 & 10 & 5 & \textbf{18} \\ 
Tennis & 6 & 4 & 2 & \textbf{12} \\ 
Volleyball & 6 & 9 & 2 & \textbf{17} \\ \hline
\textbf{Total} & \textbf{52} & \textbf{84} & \textbf{39} & \textbf{175} \\ \hline
\end{tabular}
\end{center}
\end{table}

Note that we limit the scope of study to only include games that are grounded in reality in terms of both game-play as well as visual appearance. This means that fictional or fantasy game genres have not been included in our dataset. The reason for this is two-fold. First, we use pre-trained models in our experiments that are trained on real-world data. Second, it is more difficult to define a common game genre when a fictional/fantasy game contains unique game-play elements and visual tropes not found in any other games. Any game within these genres that satisfies this grounded-in-reality criterion, however, can be considered within the scope of generalization. In future studies the scope of generalization can be expanded to cover additional game genres of interest.

\section{Game Representation Learning}
\label{sec:replearning}

This section describes the training procedure of the ConvNet visual encoder to obtain generalized representations. Since we are trying to learn general-purpose representations that can be used for a number of core AI applications in games \cite{yannakakis2018artificial} including game-playing, game content generation, affect modeling, etc., we need to formulate our training method with a proxy learning task which acts as a general-purpose learning framework. Based on the analysis carried out by Mitrovic \emph{et al.} \cite{mitrovic2020representation}, we formulate the task of learning game representations from pixels as an image classification problem.

Let's denote our game genres dataset as $\mathcal D = \{(X_i,y_i)\}$ for all pairs where $X_i$ is an RGB image of height $h$ and width $w$ drawn from the set $\mathcal X \subset \mathbb R^{h \times w \times 3}$. $y_i \in \mathcal Y$ is the game genre label for $X_i$ belonging to a set of different game genres. Thus, our image classification problem can now be defined as learning a function $f:\mathcal X \rightarrow \mathcal Y$.  
In our experiments, we are using ConvNets to estimate the function $f$ by iterating over the training dataset $\mathcal D$. Thus, the function $f$ can be seen as a composite function $f = c \circ r$ where $r:\mathcal X \rightarrow \mathbb R^d$ is estimated by the visual encoder comprising of convolutional layers, $\mathbb R^d$ is the $d$-dimensional representation space learnt by the encoder and $c: \mathbb R^d \rightarrow \mathcal Y$ is the classifier comprising of fully-connected layers giving the output class prediction. After investigating various architectures and baseline models, we select the ResNet-50 \cite{he2016deep} architecture as our visual encoder for all experiments reported in this paper due to its reasonable size ($\sim25\cdot10^6$ learnable parameters) to performance ratio. Moreover, we initialize the learnable parameters of this model with the weights learnt from pre-training it on the ImageNet dataset (available at  \textit{keras.io}). Lastly, the classifier contains two fully-connected layers with a dropout rate of $0.2$ and the last layer employs the softmax-activated cross-entropy loss function for learning class probabilities. 

In the following subsections, we first present the different data pre-processing techniques employed to prepare the input images for feeding into the neural networks. Then, we lay out the two different training approaches---Fully Supervised Learning and Supervised Contrastive Learning---that we use for learning the functions $r$ and $c$. 

\subsection{Data Pre-Processing}\label{sec:replearning_preprocessing}

Before feeding the images from our dataset to the ResNet encoder for training, we perform multiple data-preprocessing steps. First, we resize the images to $h=224$ and $w=224$ regardless of the original image dimensions. We settle upon this size for two reasons: (a) the version of pre-trained ResNet model used has been originally trained on this image size, and (b) this also happens to be the ideal image size we can fit onto our GPU hardware (8GB VRAM) for training. 

Then, we split our dataset into training set $D^T$ and validation set $D^V$ such that $D^T \cap D^V = \emptyset$ and $D^T \cup D^V = D$. Instead of naively dividing all the images in $D$ into the two sets based on their genre, we algorithmically select $D^T$ and $D^V$ such that the games selected for the training set do not overlap with those in the validation set. This ensures that the models are tested on games that are not encountered during training, enabling us to evaluate the out-of-distribution generalization \cite{arjovsky2021out} performance of our models, i.e., on new/unseen games. Moreover, this algorithm aims for roughly 75\%---25\% split and tries to pick equal ratios of games across the visual styling categories, so that the balance of \emph{retro}, \emph{modern}, and \emph{photoreal} games is maintained in both the training and validation sets. 

\begin{figure}[!tb]
\includegraphics[width=\columnwidth]{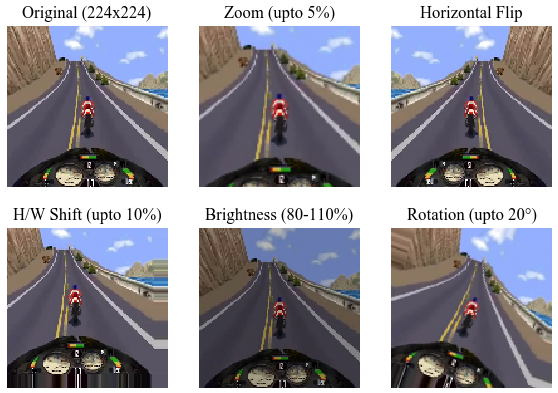}
\caption{Examples of the different image augmentation techniques used with associated random probabilities.}
\label{fig:DataAug}
\end{figure}

\begin{figure*}[!tb]
\includegraphics[width=\linewidth]{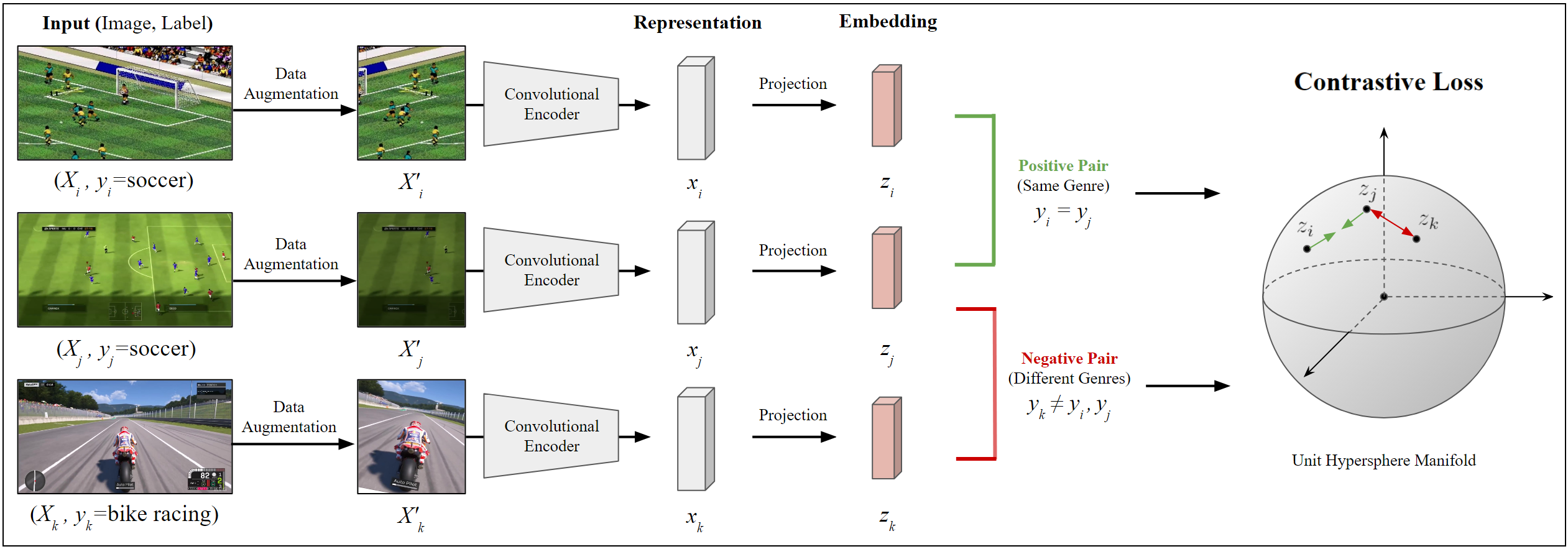}
\caption{Supervised Contrastive Learning Framework.}
\label{fig:SupConFramework}
\end{figure*}

Next, we perform image data augmentation using various techniques studied in \cite{shorten2019survey} such as \textit{horizontal flipping, zoom, brightness, height/width re-scaling} and \textit{rotation}, shown in \figurename{}~\ref{fig:DataAug}. These augmentations are re-applied with a different probability each time an image is loaded for training across multiple training epochs. This ensures that the input batches given to the neural network during each epoch are slightly different, making the training process more robust and limiting over-fitting on the training images. We shall denote this step as the function $aug: \mathbb R^{h\times w\times3} \rightarrow \mathbb R^{h\times w\times3}$ and $X^{'}_{i}=aug(X_i)$ which means $X^{'}_{i}$ is a randomly augmented version of image $X_i$. Note that this function is only applied to the images in $D^T$ and not in $D^V$.

\subsection{Fully Supervised Learning}\label{sec:replearning_surpervised}

The most popular method for training ConvNets for image classification tasks is standard supervised learning. This method trains both the encoder $r$ and classifier $c$ during the same training iteration. We sample batches of size $b=64$ from $D^T$ and for all the images $X_i$ in this batch, we obtain its augmentation $X^{'}_{i}=aug(X_i)$. Then, we compute the representation of this augmented image using the ResNet-50 encoder, given by $x_i=r(X^{'}_{i})$. The class probabilities predicted by the classifier are given by $\sigma_i=c(x_i)$ where the probabilities are normalized with softmax activation over $n=10$ classes of our dataset. The network is trained to minimize the categorical cross-entropy loss, defined as:
\begin{equation}
\label{celoss}
    \mathcal L_{ce} = - \sum_{\forall \sigma_i}y_{i}log(\sigma_i)
\end{equation}
\noindent where $y_i$ is the true label of input image $X_i$. $\mathcal L_{ce}$ is used to calculate the gradients using the Adam optimizer \cite{kingma2014adam} to update the parameters of both the encoder and the classifier networks. 
After hyper-parameter tuning, we settle on using a decaying learning-rate initialized at $0.001$, a batch size of 64 and $|D^T|/10$ training steps per iteration for this experiment. 

The results obtained from this learning method after 10 training epochs are presented in Section \ref{sec:results}. The accuracy metrics are calculated based on the class predictions; 
the silhouette scores and t-SNE embeddings are calculated using the representation $x_i$.

\subsection{Supervised Contrastive Learning}\label{sec:replearning_contrastive}

In this section, we introduce contrastive learning as an alternate method to train the ConvNet model. Contrastive learning operates with both labelled \cite{khosla2020supervised} and unlabelled \cite{chen2020simple} data; since our dataset contains game genre labels we proceed with the supervised variant and compare it to standard supervised learning in order to test and highlight the impact of this training framework in improving generalization in (game-based) computer vision tasks.

The contrastive learning framework involves training the functions $r$ and $c$ in two steps using two separate loss functions, unlike the previous method where only one loss function (Eq. \eqref{celoss}) is employed for both. The first \emph{pre-training} step for the encoder uses a pairwise loss function, namely the contrastive loss (Eq. \eqref{conloss}). In this step, we take the representation $x_i$ and use a projection network to map it onto a lower-dimensional embedding space which is a hyperspherical manifold of unit radius, as explained in \figurename{} \ref{fig:SupConFramework}. Let's denote this as $p: \mathbb R^d \rightarrow \mathbb R^{128}$ so that $z_i=p(x_i)$ gives us the embedding of image $X_i$. The function $p$ is also learnt by a fully connected neural network layer, chosen to be of size 128 in our experiments. Then, the supervised contrastive loss can be calculated on a given batch of images as:

\begin{equation}
\label{conloss}
 \mathcal L_{con} = 
 \begin{cases}
    \|{z_i - z_j}\|_2^2,& \text{when } y_i = y_j\\
    max(0,m-\|{z_i - z_j}\|_2)^2,& \text{when } y_i\neq y_j
\end{cases}
\end{equation}
 
\noindent where $m$ is the \emph{margin} hyper-parameter and its value is set to $m=1.0$ in our experiments. Eq. \eqref{conloss} is the \emph{max margin} variant of contrastive loss as proposed by Hadsell \emph{et al.} \cite{hadsell2006dimensionality}. In principle, this function pulls representations of same-class labels closer together on the hyperspherical manifold and pushes apart those that belong to different class labels, as explained in \figurename{} \ref{fig:SupConFramework} using the terms \emph{positive pair} and \emph{negative pair}. This arranges the representations so that images of the same label form a compact cluster and the clusters formed by images of different labels are as well separated from this cluster as possible. Hence, the representations learnt by the encoder under this framework are expected to be better organized in the representation space compared to the fully supervised approach. 
Preliminary experiments tested other variants of the contrastive loss function such as \emph{supervised NT-Xent} \cite{khosla2020supervised}, \emph{triplet} \cite{weinberger2006distance} and \emph{multi-class n-pairs} \cite{sohn2016improved} but none performed as well as the max margin loss.

In the second step of this framework, we train the classifier $c$ on the learned representations using the cross-entropy loss (Eq. \eqref{celoss}). The projection network $p$ is discarded at this point and the weights of the encoder $r$ are set to be non-trainable since the representations are already well-organized in the Euclidean space due to the pre-training step. At this point, the task of learning $c$ becomes trivial for the classifier. The accuracy and silhouette metrics are calculated for this method similar to the fully supervised method as described in Section \ref{sec:replearning_surpervised}.

\section{Results}\label{sec:results}
In this section, we first present an objective comparison of the two training methods in terms of classification accuracy and silhouette metrics. Then, we dive into comparing the representation spaces learned by each method and explain its importance towards evaluating generalization.
\begin{table}[!tb]
\begin{center}
\caption{Fine-tuning the pre-trained models on the \emph{Sports10} dataset after 10 epochs. We present average values across 5 runs and corresponding 95\% confidence intervals.}
\label{classres}
\begin{tabular}{|l|@{ }c@{ }|@{ }c@{ }|@{ }c@{ }|}
\hline
\textbf{\begin{tabular}[c]{@{}c@{}}Learning Method\end{tabular}} & \textbf{\begin{tabular}[c]{@{}c@{}}Training \\ Accuracy\end{tabular}} & \textbf{\begin{tabular}[c]{@{}c@{}}Validation\\  Accuracy\end{tabular}} & \textbf{\begin{tabular}[c]{@{}c@{}}Silhouette\\  Score\end{tabular}} \\ \hline
Pre-Trained (ImageNet) & - & - & \ -0.03 $\pm$ 0.01\  \\ 
Fully Supervised & \ 99.64 $\pm$ 0.08\  & \ 90.41 $\pm$ 1.53\  & \ 0.22 $\pm$ 0.01\  \\ 
Supervised Contrastive & \ 91.83 $\pm$ 0.39\  & \ 93.42 $\pm$ 0.70\  & \ 0.56 $\pm$ 0.01\  \\ \hline
\end{tabular}
\end{center}
\end{table}

\subsection{Quantitative Analysis}\label{sec:results_quantitative}

Table \ref{classres} shows the comparison of classification accuracy results between the two training methods in terms of mean and  95\% confidence interval over 5 runs with random seed initialization. We observe that while the fully supervised method achieves higher accuracy on the training data, contrastive learning achieves significantly higher accuracy ($p<0.05$) on the validation data. However, this is only a marginal improvement and overall both approaches seem to be able to learn the image classification problem. Additional experiments with different training/validation splits showed similar trends in training and testing accuracies. Merely observing the average accuracy values obtained, there does not seem to be a major advantage of using supervised contrastive learning over the conventional method. Findings appear more interesting and relevant, however, when one compares the representation spaces themselves, instead of their classification capacities. 


At the start of the training process, we use the pre-trained ResNet model with ImageNet weights. The average silhouette score for this model is $-0.03$, which means that the different game genres have poor clustering and the representations of all games are not arranged well in the Euclidean space. We observe, however, that fine-tuning this model with fully supervised training improves this average silhouette score to around $0.22$, offering an improvement in the clustering of representations by a significant margin. Fine-tuning the representation via supervised contrastive training offers a much larger (and significant) improvement that reaches silhouette scores of $0.56$, on average. This indicates that the clusters of game genres obtained via contrastive learning are much more compact and well separated compared to standard supervised learning. Based on our findings in the \emph{Sports10} dataset we conclude that supervised contrastive learning is better suited (compared to fully supervised learning) for solving domain gap problems, thereby making it the preferred method towards generalization in pixel-based game representations.

\begin{figure}[!tb]
\centerline{\includegraphics[width=0.48\textwidth]{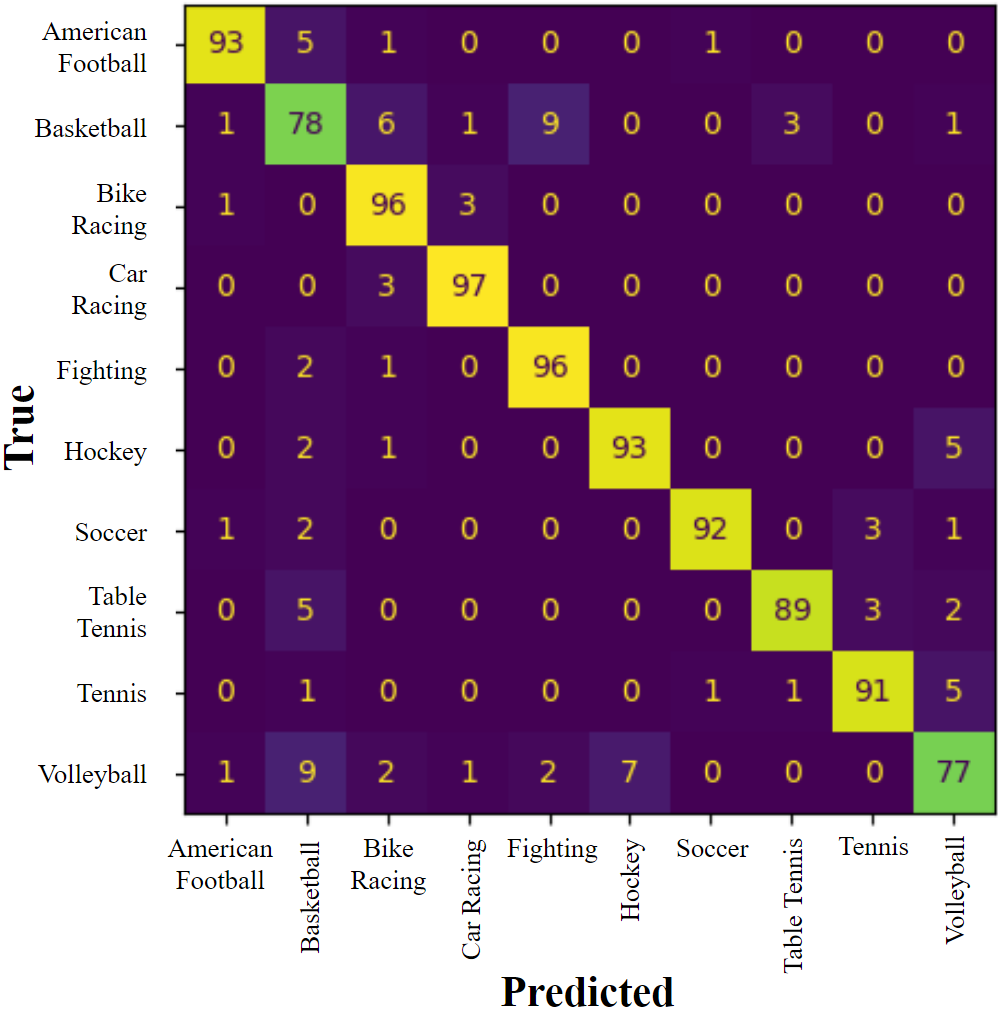}}
\caption{Confusion Matrix in terms of validation accuracy (\%).}
\label{fig:confmatrix}
\end{figure}

Lastly, we want to look into the genre-wise classification accuracy for our best performing model. \figurename{} \ref{fig:confmatrix} gives us the confusion matrix showcasing the overlap between true label and the predicted label. Based on this matrix, it seems that the two genres \textit{Volleyball} and \textit{Basketball} are tougher to classify relative to other genres. \textit{Basketball} games sometimes get misclassified as \textit{Fighting} and \textit{Bike Racing} while \textit{Volleyball} games are often confused as \textit{Basketball} or \textit{Hockey}. This happens when the representations of these genres lie closer to the clusters of other genres than their own, leading to misclassification. 

\begin{figure*}[!tb]
\begin{subfigure}{\textwidth}
  \centering
  \includegraphics[width=\linewidth]{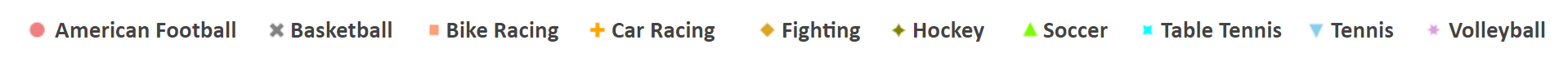}  
\end{subfigure} \\
\begin{subfigure}{.33\textwidth}
  \centering
  \includegraphics[width=\linewidth]{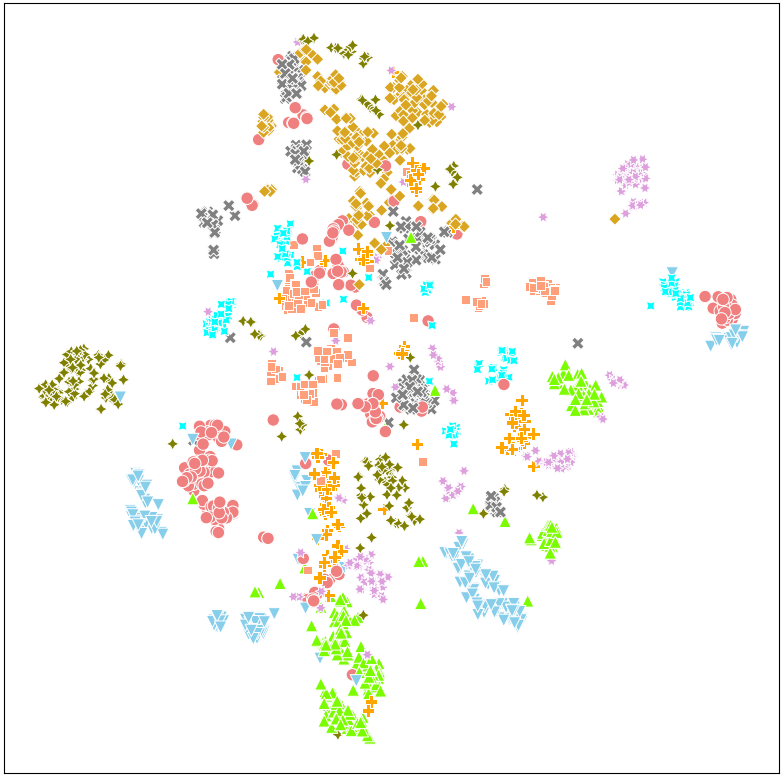}  
  \caption{ImageNet}
  \label{fig:sub-first}
\end{subfigure}
\begin{subfigure}{.33\textwidth}
  \centering
  \includegraphics[width=\linewidth]{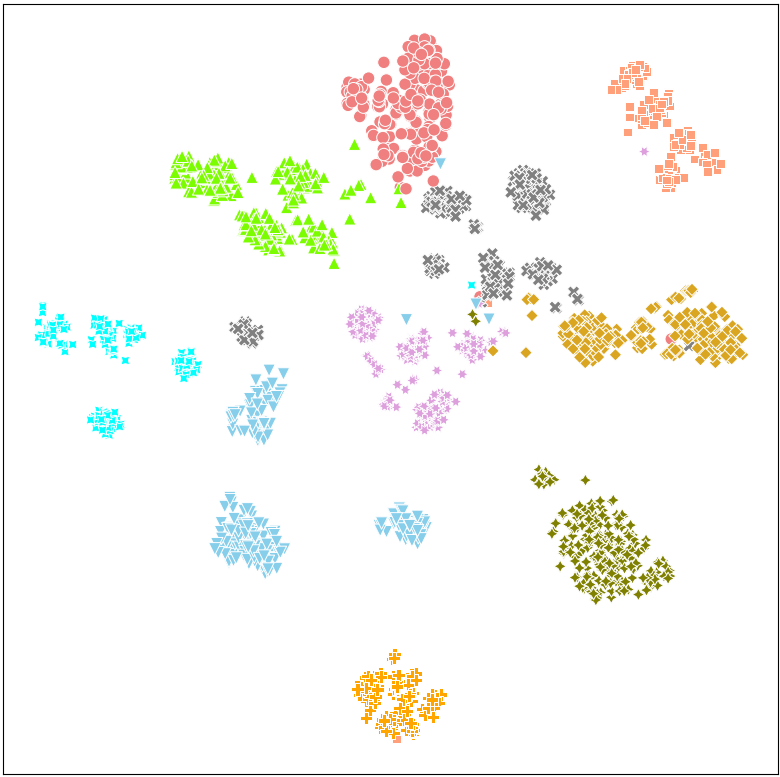}  
  \caption{Fully Supervised Learning}
  \label{fig:sub-second}
\end{subfigure}
\begin{subfigure}{.33\textwidth}
  \centering
  \includegraphics[width=\linewidth]{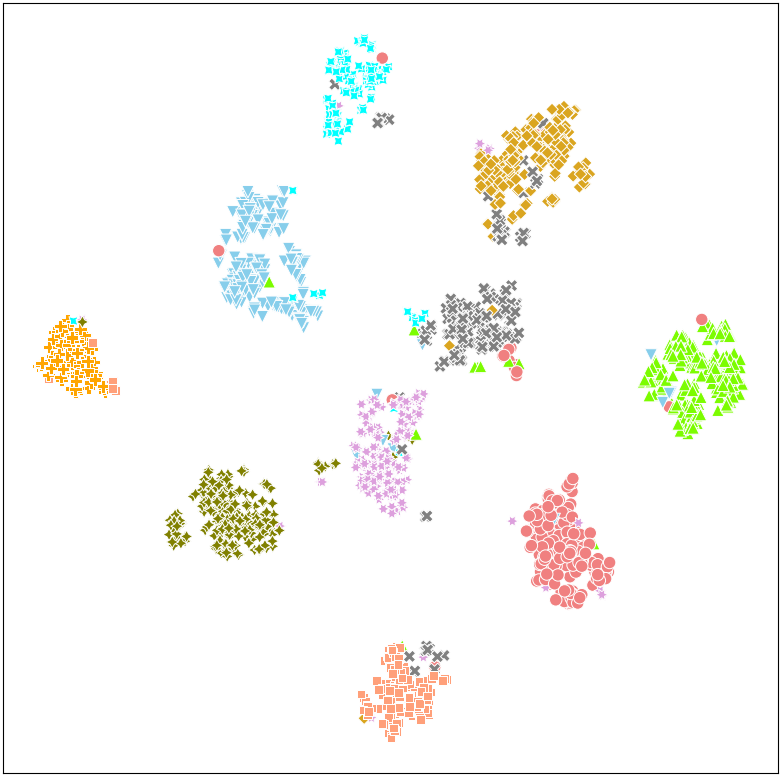}  
  \caption{Contrastive Learning}
  \label{fig:sub-third}
\end{subfigure}
\caption{Two dimensional projections of t-SNE embeddings calculated on a random sample taken from the validation set.}
\label{fig:tsne}
\end{figure*}

\subsection{Qualitative Analysis}\label{sec:results_qualitative}

In addition to the quantitative results above, we present an alternative qualitative analysis by directly visualizing the representation space and seeing the layout of the clusters for all game genres. 
\figurename{} \ref{fig:tsne} visualizes the three representation spaces: (a) ImageNet Representations without fine-tuning; (b) Fine-tuning via Fully Supervised Learning; (c) Fine-tuning via Contrastive Learning, using t-SNE on the validation dataset. Note that none of the games in the validation set were seen during training, so these results give a fair indication of how these models will perform on unseen games of these 10 genres.

Notably, we observe that the t-SNE analysis is in alignment with the silhouette scores obtained. Without training (see Fig. \ref{fig:sub-first}) all 10 game genres are poorly clustered; their clustering improves slightly when fully supervised learning is employed (see Fig.~\ref{fig:sub-second}) due to fine-tuning. The representations in this space appear to form separate clusters for each genre which are distant to other clusters. Within each cluster, however, we can observe islands of sub-clusters formed by the different games present under that genre label. The observation of Fig.~\ref{fig:sub-second} indicated that there still exist domain gaps within the genres, albeit to a lesser extent compared to the ImageNet pre-trained model. The t-SNE embeddings obtained via contrastive learning (see Fig.~\ref{fig:sub-third}) showcase the best-formed clusters, while addressing both inter-genre separability and intra-genre compactness. Such a representation space can only be expected to generalize well over any game of these genres and the applications built on top of one game are expected to be easily transferable to other games of that genre.

\section{Applications of General Representations}\label{sec:discussion}

This paper applies representation learning to a broad set of games that follow real-world patterns. Results indicate that contrastive learning can capture the genre-specific visual content while filtering out stylistic differences between games. This opens up a number of interesting directions for applications that benefit from general pixel-based representations.

An obvious application for generalized pixel-based representations is game playing agents. For example, the current learned representation can be used with the Google Research Football Environment \cite{kurach2019google} to initialize the visual encoder of an imitation- or reinforcement-learning agent. With most genre-specific information (such as pattern of football pitch, goalposts) already present in these representations, fine-tuning the task-specific visual information (such as position of players, ball) or learning the control-policy becomes much more sample efficient compared to starting from scratch.


Procedural content generation (PCG) tends to use game-specific representations (e.g. tilesets) to produce content (e.g. levels). Therefore, pixel-based general representations are more suited for evaluating the content rather than for explicitly generating it. As an evaluation mechanism, general representations can be very beneficial to e.g. evaluating the typicality \cite{ritchie2007criteria} of a generated game compared to similar examples of its genre. Moreover, the general representations can serve for coherence evaluation when combining generators of multiple facets (e.g. visuals, level structures, rules) in order to assess whether the resulting game fits the patterns of a specific genre. This would allow orchestration of game content \cite{liapis2019orchestrating} not only to avoid nonsensical combinations but also to identify the genre of the generated game and potentially create games of different genres within one run.

General representations also seem ideally suited for PCG via Machine Learning \cite{summerville2018pcgml}. Guzdial \emph{et al.} \cite{guzdial2016videos} investigate how gameplay videos can be mined for level patterns. Contrastive learning can augment this line of research by detecting general level patterns across multiple games of the same genre and thus generate content for any game of this genre. It should be noted that the tested sports dataset has fairly uniform levels per genre (e.g. the same football pitch in all football games) and thus the level patterns are less critical in terms of content detected. Future work should explore to which extent contrastive learning can be used to detect level patterns in more mechanically diverse games within e.g. the platformer genre. 

Another important application is player modeling \cite{yannakakis2013player, yannakakis2018artificial}. Work on identifying highlights in gameplay videos \cite{ringer2018unsupervised} could benefit from concise, genre-specific representations for e.g. classification purposes. Moreover, such general representations can be used to model affective responses based on game footage alone, extending current work \cite{makantasis2019pixels,makantasis2021pixels} that trains custom ConvNets for each game. As larger and more mixed-genre datasets for players' affect become available \cite{melhart2021again}, applying contrastive learning for pre-processing to remove stylistic differences will be crucial. 

Finally, the general representations can be utilized to learn game dynamics with pixel-based forward models. GameGAN \cite{kim2020learning} is a good example of learning a physics engine of \textit{Pac-Man} (Namco 1980) that predicts future game states at pixel-level based on user inputs. Such forward models could be shared across different games of the same genre if they are built using generalized representations. This can massively reduce the workload of creating new game state representations (as well as gameplaying agents) for genres that already have a neural engine.
    

\section{Conclusion}\label{sec:conclusion}

In this study we introduced the use of contrastive learning as a method for yielding general game representations with the aim of improving their reusability across different games without requiring re-training. We test our hypothesis that contrastive learning is beneficial for pixel-based representation learning on a new dataset, named \emph{Sports10}, that contains 100k labelled images from 175 games across 10 sports genres. 
Our experimental results suggest that contastive learning outperforms conventional supervised learning in classifying game genres. More importantly for the purposes of this paper, the representations learned via contrastive learning yield more compact clusters of game representations belonging to the same genre which are, in turn, better separated from clusters of other genres. It appears that learning through contrasting game images leads to fewer domain gap issues compared to the representations learned under conventional supervised learning. The methods and results of the paper form a basis for further research on areas of game AI \cite{yannakakis2018artificial}: from gameplaying agents and game world models, all the way to pixel-based procedural content generation and player modeling.

\bibliography{genre_contrastive_learning}

\end{document}